\pdfoutput=1

\documentclass[11pt]{article}

\usepackage{acl}

\usepackage{times}
\usepackage{latexsym}

\usepackage[T1]{fontenc}

\usepackage[utf8]{inputenc}

\usepackage{microtype}

\usepackage{inconsolata}

\usepackage{color,xcolor}

\usepackage{pifont}
\usepackage{bm}
\usepackage{amsfonts}

\usepackage{graphicx}
\usepackage[normalem]{ulem}
\usepackage{multirow}
\usepackage{amsmath}
\usepackage{url}

\usepackage{tikz}
\usepackage{xcolor}
\usepackage{tcolorbox}
\usepackage{color,xcolor}

\usepackage{amssymb}
\usepackage{amsopn}
\usepackage{graphicx}
\usepackage{multirow}
\usepackage[english]{babel}
\usepackage{bm}
\usepackage{array}
\usepackage{setspace}
\usepackage{todonotes}

\usepackage{xspace}
\usepackage{amsfonts}
\usepackage{array,multirow}
\usepackage{pgfplots}
\usepackage{tikz}
\usepackage{verbatim}
\usepackage{subfig}
\usepackage{arydshln}
\usepackage{booktabs}

\usepackage{hyperref}
\usepackage{multirow}
\usepackage{setspace}
\usepackage{booktabs}
\usepackage{graphicx} 
\usepackage{arydshln}
\usepackage{url}
\usepackage{CJKutf8}

\title{Improving Machine Translation with Large Language Models: A Preliminary Study with Cooperative Decoding}

\author{Jiali Zeng, \
Fandong Meng, \ Yongjing Yin, \ Jie Zhou \\
Pattern Recognition Center, WeChat AI, Tencent Inc \\
\texttt{\{lemonzeng,fandongmeng,yongjingyin,withtomzhou\}@tencent.com} \\
}

\begin{document}
\begin{CJK}{UTF8}{gbsn}

\maketitle
\begin{abstract}


Contemporary translation engines based on the encoder-decoder framework have made significant strides in development.
However, the emergence of Large Language Models (LLMs) has disrupted their position by presenting the potential for achieving superior translation quality.
To uncover {\it the circumstances in which LLMs excel and explore how their strengths can be harnessed to enhance translation quality},
we first conduct a comprehensive analysis to assess the strengths and limitations of various commercial NMT systems and MT-oriented LLMs. 
Our findings indicate that neither NMT nor MT-oriented LLMs alone can effectively address all the translation issues, but MT-oriented LLMs show promise as a complementary solution to NMT systems.
Building upon these insights, we 
propose {\bf Co}operative {\bf Dec}oding (CoDec), which treats NMT systems as a pretranslation model and MT-oriented LLMs as a supplemental solution to handle complex scenarios beyond the capability of NMT alone.
Experimental results on the WMT22 test sets and a newly collected test set WebCrawl demonstrate the effectiveness and efficiency of CoDec, highlighting its potential as a robust solution for combining NMT systems with MT-oriented LLMs in the field of machine translation.

\end{abstract}

\section{Introduction}


\begin{figure}[!t]
\centering
\includegraphics[width=1.0\linewidth]{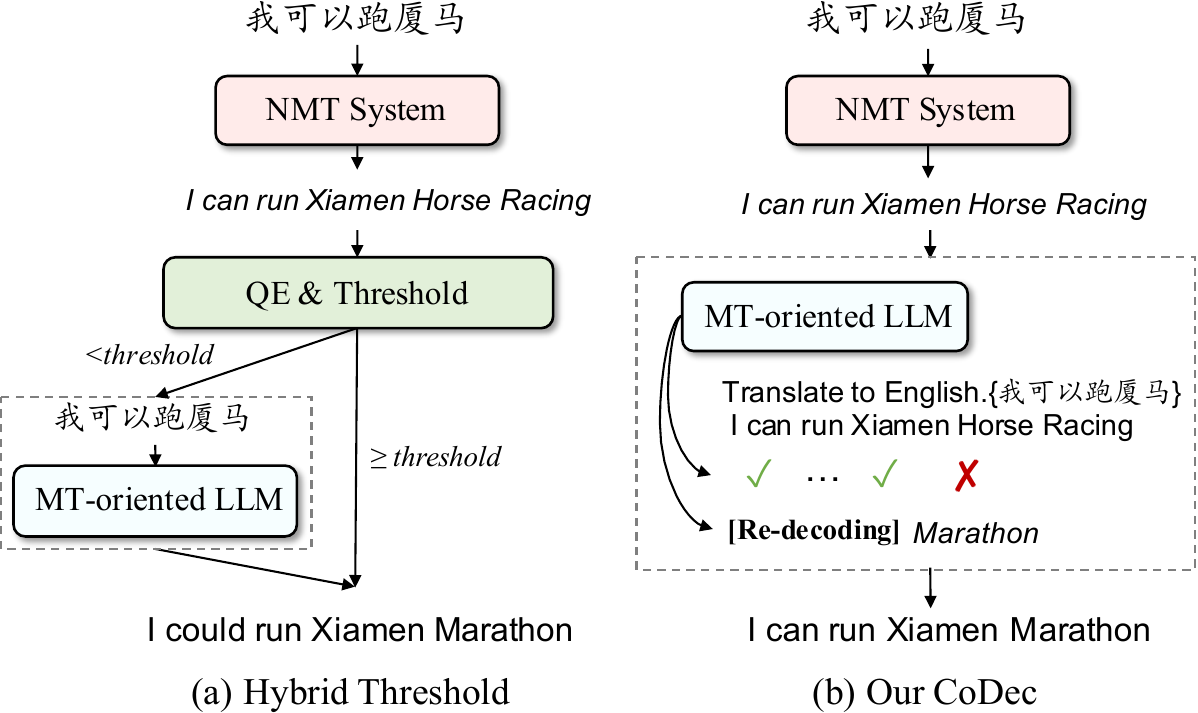}
\caption{
{\bf Comparison between the Hybrid Threshold and CoDec frameworks}.
CoDec is more efficient than Hybrid Threshold as it eliminates the need for an extra quality evaluation module and autoregressive generation of the whole translation using MT-oriented LLM.
}
\label{fig_inro}
\end{figure}


Over the years, the encoder-decoder framework has established Neural Machine Translation (NMT) models as the prevailing standard, achieving impressive translation quality through extensive training on large-scale and high-quality parallel data \citep{Vaswani:NeurIPS2017,freitag-firat-2020-complete,fan-etal-2021-beyond}. 
Commercial machine translation engines, e.g., Google Translate, are proficient in addressing the majority of translation requirements.
More recently, with the emergence of generative large language models (LLMs), the position of traditional NMT models has been challenged \citep{gpt3,arxiv2023:openai_gpt4}. 
While commercial LLMs like OpenAI's GPT-4 currently perform well in translation \citep{Hendy,Zhu,emnlp/LinMAWCSOGBDPSK22,Agrawal}, they are constrained by their interface nature, thereby limiting further customization and improvement due to privacy concerns in industrial applications.
A more promising approach involves fine-tuning relatively smaller LLMs (i.e., fewer than 13B parameters) to create LLMs specifically tailored for MT \citep{zeng-etal-2023-tim, Arxiv2023:bayling, Jiao_ParroT}. 

In this context,
this study aims to investigate the following research questions:
{\it 
In which scenarios do MT-oriented LLMs demonstrate superior performance to conventional NMT models, and how can we leverage the strengths of the two paradigms to enhance translation quality?}

To begin, we conduct a comprehensive analysis
into the characteristics of translations generated by commercial NMT systems and MT-oriented LLMs. 
Our findings reveal that commercial NMT systems excel at producing adequate translations in specific domains or languages. 
Conversely, MT-oriented LLMs demonstrate proficiency in generating authentic-sounding translations and handling infrequent words that are not effectively processed by NMT systems. 
In summary, MT-oriented LLMs can serve as valuable fallback systems in cases where the output of commercial NMT systems is unsatisfactory.

To complement NMT with MT-oriented LLMs, 
\citet{Hendy} introduced the Hybrid Threshold approach (Figure \ref{fig_inro}(a)), which employs the NMT system as the primary translation system.
When the translation fails to meet the quality threshold determined by the quality estimation (QE) module, an alternative translation is generated using a GPT-like model.
However, this approach faces two primary challenges.
First, existing reference-free metrics struggle to align with human judgment,
resulting in inaccuracies being propagated \citep{freitag-etal-2021-results,freitag-etal-2020-bleu,ma-etal-2019-results,rei-etal-2022-comet}. Second, the integration of neural quality estimation modules and the sequential execution by LLMs leads to increased decoding time \citep{tay-etal-2023-efficient,xu-etal-2021-asurvey}, which poses concerns for efficient translation in practical applications.

To address the above issues, 
we propose an efficient implementation approach for system ensembles called {\bf Co}operative {\bf Dec}oding (CoDec). 
As illustrated in Figure \ref{fig_inro}(b), 
the NMT system functions as the front-end module, generating an initial translation draft for a given input sentence. 
Subsequently, the MT-oriented LLM serves as both an evaluator and a refiner, which firstly evaluates the draft from a language modeling perspective, and then the LLM refines the partial translation starting from a specific position where the token in the draft is not among the top-$k$ token candidates suggested by the LLM.
Since the evaluation process takes advantage of parallel computation and the front-end module can handle most situations effectively, CoDec is more efficient compared to using LLMs for complete decoding.

The contributions of this paper are three-fold:
\begin{itemize}
    \item We conduct in-depth analyses on the WMT22 test sets and a newly collected test set, WebCrawl, to identify the strengths and weaknesses of traditional NMT systems and MT-oriented LLMs, finding that MT-oriented LLMs can complement NMT systems.
    \item We present CoDec, a novel hybrid framework that synergizes the strengths of NMT systems and MT-oriented LLMs.
    By harnessing the complementary capabilities of MT-oriented LLMs, CoDec effectively overcomes the limitations of traditional NMT systems\footnote{We release code and the translations of different systems at https://github.com/lemon0830/CoDec}.
    \item 
    We evaluate the performance of CoDec on various test sets.
    Our CoDec, without the need for an additional quality estimation module, achieves competitive or even better performance than Hybrid Threshold. 
    Furthermore, CoDec offers a significant acceleration advantage, achieving an acceleration ratio of approximately 2x compared to directly using LLM for generation. 
\end{itemize}


\section{Related Work}
\subsection{Large Language Models on Machine Translation}
Research on Large Language Models (LLMs) for machine translation can be broadly divided into two categories: utilizing LLMs as an interface and optimizing them for specific translation tasks.
For the former, \citet{Hendy} evaluate ChatGPT, GPT3.5, and text-adavinci-002 in eighteen translation directions, while \citet{Zhu} assess XGLM, BLOOMZ, OPT, and ChatGPT across 202 directions and 102 languages.
Other researchers explore strategies for selecting translation exemplars \citep{emnlp/LinMAWCSOGBDPSK22,Agrawal} and incorporating external knowledge \citep{arxiv2023:chainofdict} to enhance GPT translation.
Fine-tuning smaller models (e.g., 7B) specifically for translation tasks has attracted increasing attention \citep{zeng-etal-2023-tim,Arxiv2023:bayling, Jiao_ParroT}.
Diverging from existing approaches, our research focuses on examining the capabilities and limitations of commercial NMT systems and MT-oriented LLMs and developing efficient hybrid frameworks that leverage their respective strengths.

\begin{table*}[!t]
\centering
\small
\setlength{\tabcolsep}{2mm}{
\begin{tabular}{l|cccc|cccc}
\toprule
{\bf System} & {\bf COMET} & {\bf COMETk.} & {\bf COMET} & {\bf COMETk.} & {\bf COMET} & {\bf COMETk.} & {\bf COMET} & {\bf COMETk.} \\
\midrule
& \multicolumn{2}{c}{\it DE$\Rightarrow$EN} & \multicolumn{2}{c|}{\it EN$\Rightarrow$DE} & \multicolumn{2}{c}{\it ZH$\Rightarrow$EN} & \multicolumn{2}{c}{\it EN$\Rightarrow$ZH} \\
WMT-Best & 85.0 & 81.4 & 87.2 & 83.6 & 81.0 & 77.7 & 86.7 & 82.0 \\
\hdashline[0.5pt/0.5pt]
GoogleMT & {\bf 85.8} & {\bf 81.8} & {\bf 88.1} & {\bf 84.1} & {\bf 82.7} & {\bf 79.3}  & {\bf 88.2} & {\bf 82.7} \\
MicroMT & 85.1 & 81.4 & 87.4 & 83.7 & 80.3 & 77.5 & 86.0 & 81.3 \\
\hdashline[0.5pt/0.5pt]
BayLing-7B & 83.2 & 80.1 & 82.1 & 79.2 & 77.5 & 75.1 & 84.4 & 79.6 \\
TIM-13B & 84.4 & 81.0 & 86.4 & 83.1 & 80.8 & 77.8 & 87.6 & 82.3 \\
\midrule
& \multicolumn{2}{c}{\it RU$\Rightarrow$EN} & \multicolumn{2}{c|}{\it EN$\Rightarrow$RU} & \multicolumn{2}{c}{\it JA$\Rightarrow$EN} & \multicolumn{2}{c}{\it EN$\Rightarrow$JA} \\
WMT-Best & 86.0 & 81.7 & {\bf 89.5} & {\bf 84.4} & 81.6 & 80.3 & 89.3 & 85.8 \\
\hdashline[0.5pt/0.5pt]
GoogleMT & {\bf 86.6} & {\bf 82.0} & {\bf 89.5} & 84.2 & {\bf 84.0} & {\bf 81.7} & {\bf 90.2} & {\bf 86.5} \\
MicroMT & 85.5 & 81.1 & 88.7 & 83.6 & 81.5 & 80.1 & 88.0 & 85.3 \\
\hdashline[0.5pt/0.5pt]
BayLing-7B & 82.5 & 79.3 & 74.7 & 70.6 & 72.2 & 72.5 & 71.2 & 73.5 \\
TIM-13B & 84.2 & 80.8 & 86.7 & 82.5 & 80.8 & 79.8 & 87.5 & 84.5 \\
\bottomrule
\end{tabular}}
\caption{
\label{tab_results_main_result}
{\bf Experimental results on the WMT22 test sets}. 
MT-oriented LLMs have the potential to achieve comparable performance to
commercial NMT systems, eliminating the need for rule-based engineering techniques.
}
\end{table*}

\subsection{Accelerate Generation for Large Language Models}

Efforts to improve the inference efficiency of LLMs have been ongoing for several years \citep{tay-etal-2023-efficient,xu-etal-2021-asurvey}, leveraging techniques such as knowledge distillation \citep{hinton-etal-2015-distilling,jiao-etal-2020-tinybert,wang-etal-2020-minilm}, quantization \citep{shen-etal-2020-qbert,sun-etal-2020-mobilebert}, pruning\citep{fang-etal-2020-reducing}, and others \citep{kim-etal-2021-length,lei-2021-attention}.
The most related work is to leverage speculative execution \citep{burton-1985-speculative, john-david-2012-computer} for the speedup of autoregressive models.
\citet{stern-etal-2018-blockwise} propose to decode several tokens in parallel to accelerate greedy decoding.
For LLMs, speculative decoding \citep{chen-etal-2023-accelerating,Leviathan-etal-2023-fast} uses an additional draft model and generates sequences with sampling.
\citet{yang-etal-2023-inference} copy some tokens from retrieved reference text to the decoder, which are validated with output probabilities.
\citet{santilli-etal-2023-accelerating} reframe MT's standard greedy autoregressive decoding procedure with a parallel formulation.
We are pioneers in using speculative execution as a fusion approach for commercial NMT systems and MT-oriented LLMs, without requiring an auxiliary quality estimation module or modifications to the target LLMs' parameters.

\section{Preliminary Experiments} \label{sec_pre_exp}

In this section, we conduct a series of analyses to quantitatively investigate 
{\it the characteristics of translations from different systems}. 

\subsection{Setup}
\paragraph{Commercial NMT Systems \& MT-oriented LLMs.}
Our focus is the use of MT-oriented LLMs in industrial settings, and the chosen commercial NMT systems consist of Google Translate ({\it GoogleMT} for brevity)\footnote{https://translate.google.com/} and Microsoft Translate ({\it MicroMT} for brevity)\footnote{https://www.bing.com/translator} due to their strong performance and high reproducibility.

Regarding MT-oriented LLMs, we utilize {\it BayLing-7B} \citep{Arxiv2023:bayling}.
We directly use the translations released on GitHub\footnote{https://github.com/ictnlp/BayLing/tree/main/exp/translation \_benchmark/bayling-7b}.
Additionally, we develop an in-house MT-oriented LLM, trained on human-written validation data from previous WMT competitions\footnote{https://www.statmt.org/wmt22/translation-task.html}, such as the newstest2017-2021 of German$\Leftrightarrow$English, Chinese$\Leftrightarrow$English, Russian$\Leftrightarrow$English, and Jappanese$\Leftrightarrow$English.
In addition, we have incorporated high-quality bilingual sentence pairs in Chinese$\Leftrightarrow$English, German$\Leftrightarrow$English, and Russian$\Leftrightarrow$English, resulting in a total of two million sentences in our training data. 
According to the data license of WMT22, the data released for the General MT task can be freely used for research purposes.
We fine-tune the tigerbot-13b-base\footnote{https://huggingface.co/TigerResearch/tigerbot-13b-base} with TIM \citep{zeng-etal-2023-tim} as the final MT-oriented LLM\footnote{https://huggingface.co/Lemoooon/TIM\_13B\_forCoDec}.

\paragraph{Automated MT Metrics.}
We follow previous studies \cite{Hendy,Zhu,zeng-etal-2023-tim, Arxiv2023:bayling} to utilize COMET-22 (wmt22-COMET-da) \citep{rei-etal-2021-cometqe}, and COMETkiwi (wmt22-COMET-kiwi-da) \citep{rei-etal-2022-comet22} for reference-free quality estimation. 
We also report ChrF \citep{popovic-etal-chrF} and SacreBLEU \citep{papineni2002bleu:} in Table \ref{tab_results_chrf_bleu} in Appendix.

\subsection{Analyses on WMT22 test sets} \label{sec_wmt22}
To prevent data leakage \citep{arxiv2023:unreasonable}, we analyze the WMT22 test sets.
Detailed statistics are reported in Appendix \ref{appendix_wmt22}.

\begin{figure*}[!t]
\centering
\includegraphics[width=1.0\linewidth]{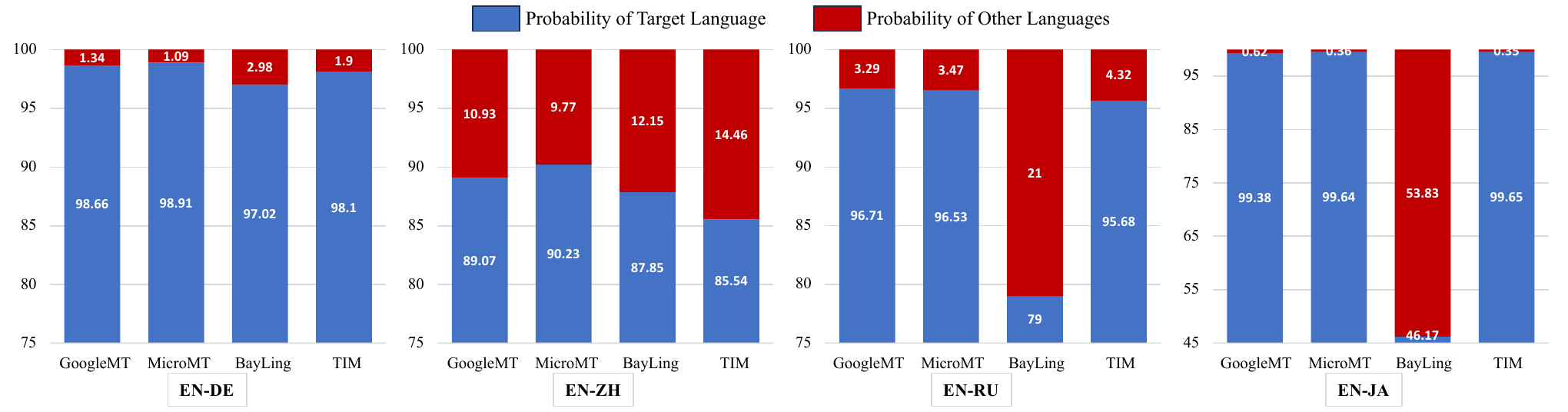}
\caption{
{\bf Off-target rates (\%) of translations.} MT-oriented LLMs (i.e., BayLing and TIM) exhibit a higher prevalence of off-target translations than NMT systems.
}
\label{fig_lang_prob_en2many}
\end{figure*}

\begin{figure*}[!t]
\centering
\includegraphics[width=1.0\linewidth]{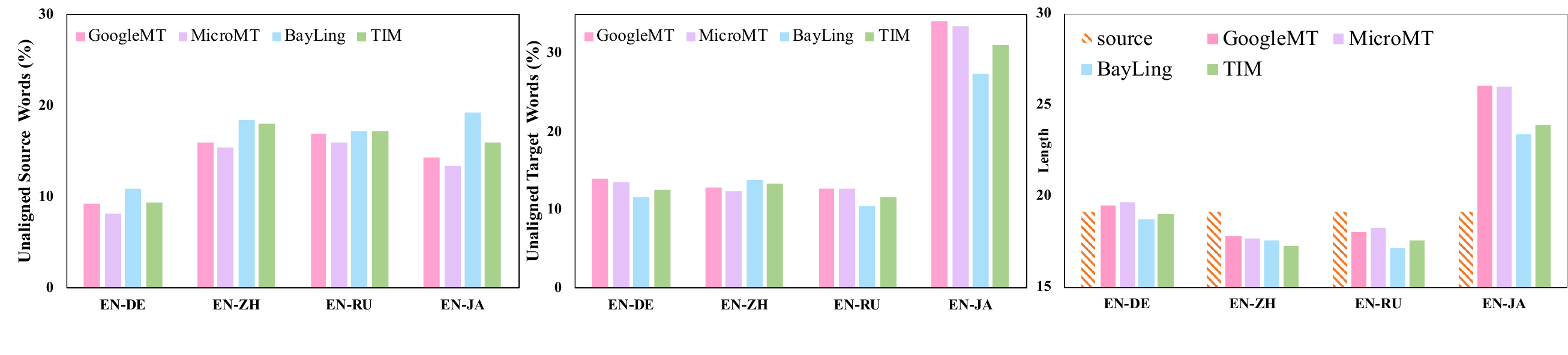}
\caption{
{\bf Comparison of unaligned source words, unaligned target words, and the length of translations.} 
MT-oriented LLMs consistently generate translations that are noticeably shorter in length and have a higher occurrence of unaligned source words across the test sets when compared to NMT models.
\label{fig_unaligned}
}
\end{figure*}

\paragraph{Main Results.} 
The experimental results are illustrated in Table \ref{tab_results_main_result}. 
We have made the following observations:
1) {\it GoogleMT} and {\it MicroMT} showcase excellent performance.
They consistently outperform the {\it WMT winner} in most of the language pairs, highlighting the robust capabilities of these well-established translation engines.
2) Despite the existing performance gap,
MT-oriented LLMs still have untapped potential for further improvement. 
Notably, {\it TIM} outperforms {\it BayLing} by a significant margin across all language pairs. 
Moreover, {\it TIM} exhibits slightly inferior performance compared to {\it MicroMT} on most test sets.
This suggests that employing more effective fine-tuning methods with large amounts of high-quality parallel data can enhance
the translation capabilities of MT-oriented LLMs, making them close to commercial NMT systems.

\paragraph{Off-target Rates.}

Off-target indicates translations generated by machines involve segments of wrong languages or code-mixing, presenting a significant challenge in multilingual neural machine translation \citep{chen-etal-2023-target,zhang-etal-2020-improving}. 
Here, we use langdetect\footnote{https://github.com/Mimino666/langdetect} to identify the language of each translation. 
The off-target rate of a translation is the subtraction of the probability of the target language prediction from 1.
For a test set, we compute the average off-target rate across all the sentences.

As depicted in Figure \ref{fig_lang_prob_en2many}\footnote{Due to limited space, we only present the results for English-to-Many translations here. The results for Many-to-English can be found in Figure \ref{fig_lang_prob} in Appendix.}, 
the MT-oriented LLMs tend to produce translations with higher off-target rates compared to NMT systems.
Specifically, {\it BayLing} exhibits off-target rates of 21\% and 53.83\% for EN$\Rightarrow$RU and EN$\Rightarrow$JA translations, respectively, which falls outside the language scope covered by the training data.
This highlights a more pronounced off-target issue in LLMs, especially in zero-shot scenarios.
In contrast, {\it TIM} achieves notably lower off-target rates in EN$\Rightarrow$RU and EN$\Rightarrow$JA compared to {\it BayLing}.
We speculate that this can be attributed to TIM's incorporation of corresponding training data, which enhances its ability to handle language switching and produce more accurate translations.

\begin{figure*}[!t]
\centering
\includegraphics[width=0.95\linewidth]{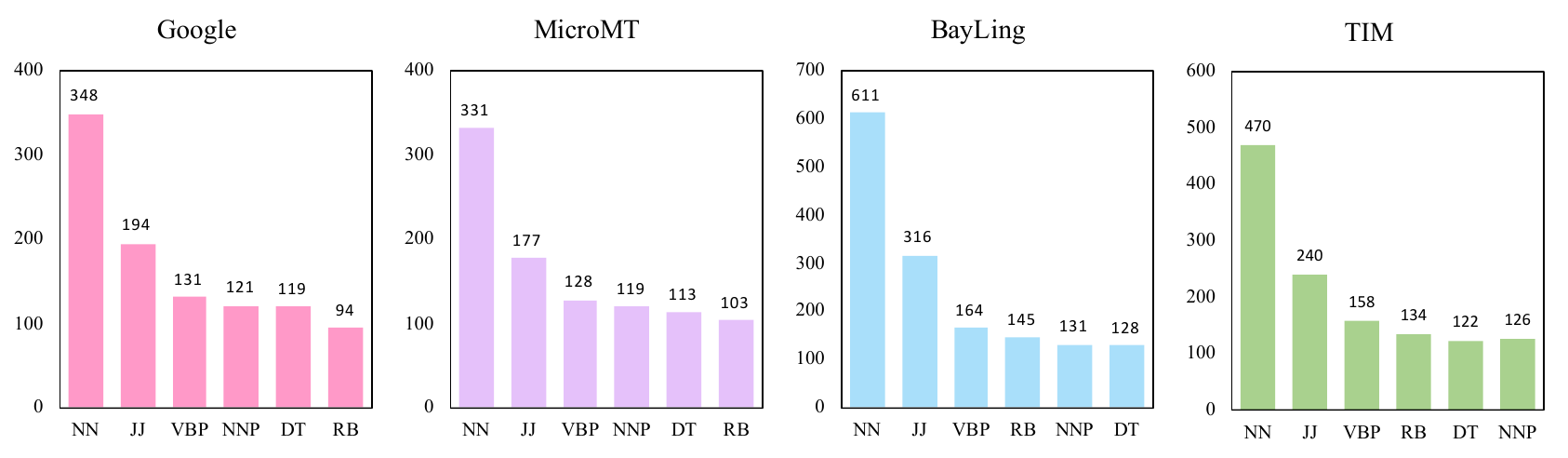}
\caption{
{\bf POS tags of unaligned source words.}
We show the top-6 POS tags of USW on the WMT22 EN$\Rightarrow$ZH test set, and the incremental USW of MT-oriented LLMs mainly lies in nouns (NN) and adjectives (JJ).
\label{fig_pos}
}
\end{figure*}

\begin{figure*}[!t]
\centering
\includegraphics[width=1.0\linewidth]{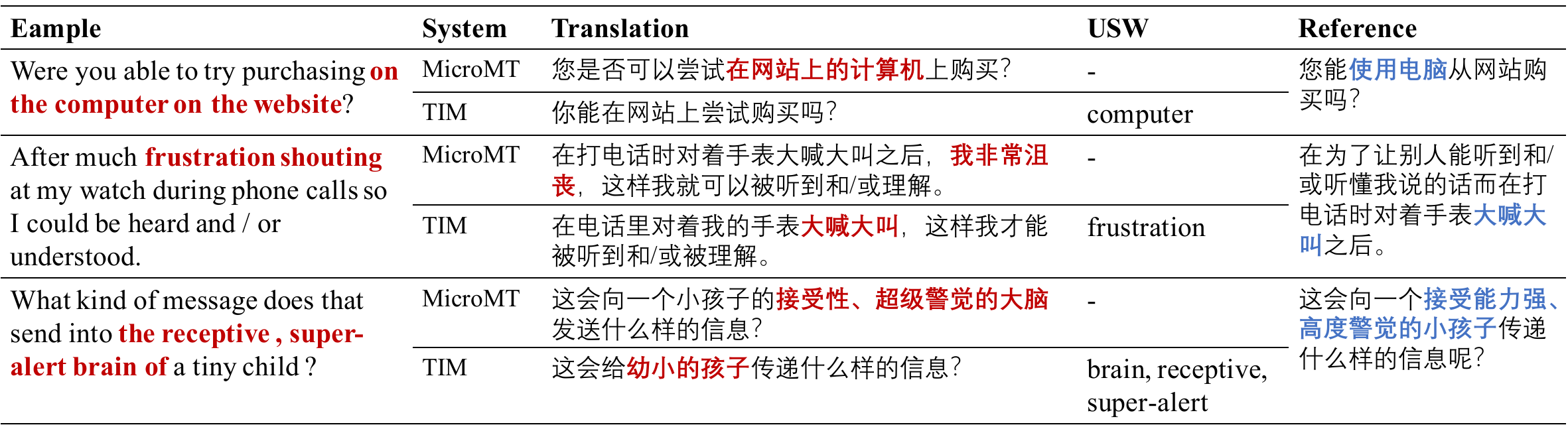}
\caption{
{\bf Examples of free translation generated by MT-oriented LLM.}
MT-oriented LLMs often produce shorter translations with significant paraphrasing, maintaining the original meaning while using different words and sentence structures.
\label{fig_case_study}
}
\end{figure*}

\paragraph{Unaligned Source/Target Words.}
To assess the literalness of the translation, we follow \citet{raunak-etal-2023-gpts, Hendy} to calculate the number of source and target words that do not align on a word-to-word basis.
More details can be found in Appendix \ref{appendix_usw}.
The left portion of Figure \ref{fig_unaligned} illustrates that the MT-oriented LLMs incur a notably larger number of unaligned source words across the test sets than the NMT counterpart.

We examine the top six part-of-speech (POS) tags by NLTK toolkit \cite{bird-etal-2009-nltk} of the unaligned source words (in Figure \ref{fig_pos}). 
The difference mainly lies in nouns (NN) and adjectives (JJ), 
indicating the possibility of increased paraphrasing or a higher degree of inadequacy, such as omitted or inserted content.
However, back to the middle part of Figure \ref{fig_unaligned}, the number of unaligned target words of the MT-oriented LLMs does not significantly differ from those of NMT systems, 
suggesting that the adequacy of translations produced by LLMs is comparable to NMT.

Additionally, we calculate the average word count in the generated translations. As depicted in Figure \ref{fig_unaligned}, MT-oriented LLMs tend to produce shorter sentences, utilizing concise and precise language.
Humans often use concise language, especially in conversations, which is abundant in the training corpus of LLMs. This influence may result in LLMs generating shorter translations.

Figure \ref{fig_case_study} presents several examples that highlight translation differences. For instance, the phrase ``frustration shouting'' should be translated as ``大喊大叫 (scream in frustration)''. While {\it MicroMT} aims for fidelity by using translation augmentation segments like ``我非常沮丧 (I feel extremely frustrated)'', {\it TIM} demonstrates a better understanding of the entire sentence and provides more accurate translations. 
However, in the third example, {\it TIM} overlooks the inclusion of the expression ``the receptive, super-alert brain of a tiny child'' from the source text, resulting in a certain degree of translation oversight. 
In summary, MT-oriented LLMs tend to generate shorter translations with substantial paraphrasing, where the original text is rephrased using different words and sentence structures while preserving the same meaning.


\subsection{Analyses on Web Crawl test sets} \label{sec_webcrawl}

The WMT22 test set is meticulously screened and annotated, with source sentences free of errors and from common domains. While the sentences have strong syntactic structures and grammatical correctness, real-world translation scenarios may not always have these ideal conditions. 
To reflect practical challenges, we collected a challenging test set from the open domain through web crawling.
Here, we focus on Chinese$\Leftrightarrow$English directions.
To acquire the data, we follow the process outlined in Appendix \ref{appendix_webcrawl}.

\begin{table*}[!t]
\centering
\small
\setlength{\tabcolsep}{0.6mm}{
\begin{tabular}{ll}
\toprule
{\bf System} & {\bf Translation} \\
\midrule
\multicolumn{2}{c}{\it Terminology/abbreviations} \\
Source & 刚好准备去厦门旅游，还能顺便跑个厦马, \\
GoogleMT & I was just about to go to Xiamen for a trip, and I could also run the Xiamen Horse Racing. \\
TIM & Just ready to go to Xiamen for a trip, and can also run the Xiamen Marathon, \\
\midrule
\multicolumn{2}{c}{\it Ill-informed text} \\
Source & Use a no. 6 ﬁ lbert to create the illusion of the rungs and \ \ the back of the chair on the left. \\
GoogleMT & 使用否。 6 榛子创建左边的梯级和椅背的幻觉。 \\
TIM & 用 6 号画笔在左边的椅子上创造梯级和椅背的错觉。 \\
\midrule
\multicolumn{2}{c}{\it Complex，Repetition-containing} \\
Source & 拯救剧荒$｜$《爱之全蚀》啊啊啊啊啊啊啊两个亲亲怪！！ \\
GoogleMT & Save the drama$｜$``Total Eclipse of Love'' Ahhhhhhhhhhhhhhhhhhhhhhhhhhhhhhhhhhhhhhhhhhhhhhhhhhh\\
& hhhhhhhhhhhhhhhhhhhhhhhhhhhhhhhhhhhhhhhhhhhhhhhhhhhhhh being to To to to toto to Eclips e Love! ! \\
TIM & Save the drama drought $｜$ ``Total Eclipse of Love'' ah ah ah ah ah two kissing monsters!! \\
\bottomrule
\end{tabular}}
\caption{
{\bf Case Study}. We present examples of several translation challenges that pose difficulties for NMT systems but are effectively mitigated by MT-oriented LLMs.
}
\label{tab_case_study}
\end{table*}

\begin{table}[!t]
\centering
\small
\begin{tabular}{lccc}
\toprule
{\bf System} & {\bf GoogleMT} & {\bf MicroMT} & {\bf TIM} \\
\midrule
\multicolumn{4}{c}{\it ZH$\Rightarrow$EN} \\
COMET$\uparrow$ & 64.4 & 59.2 & \bf{65.1} \\
COMETk.$\uparrow$ & 59.1 & 57.4 & \bf{61.9} \\
{\#Length} & 56.81 & 51.38 & 52.09 \\
{Off-Target$\downarrow$} & 1.08\% & 1.04\% & 1.82\% \\
USW & 13.87\% & 13.70\% & 16.52\% \\
UTW & 31.29\% & 25.08\% & 27.91\% \\
\midrule
\multicolumn{4}{c}{\it EN$\Rightarrow$ZH} \\
COMET$\uparrow$ & 71.2 & 68.9 & \bf{74.6} \\
COMETk.$\uparrow$ & 60.5 & 62.9 & \bf{64.9} \\
{\#Length} & 48.51 & 47.99 & 46.57 \\
{Off-Target$\downarrow$} & 15.55\% & 14.08\% & 22.41\% \\
USW & 21.76\% & 20.23\% & 25.70\% \\
UTW & 16.55\% & 13.64\% & 16.39\% \\
\bottomrule
\end{tabular}
\caption{
\label{tab_results_webcrawl}
{\bf Experimental results on WebCrawl test sets}. 
LLMs hold promise as potential fallback systems when NMT systems fail to meet quality expectations.
}
\end{table}

\paragraph{Main Results.} 
Similarly, we compute various evaluation metrics for NMT systems and TIM on the WebCrawl test sets.
As shown in Table \ref{tab_results_webcrawl}, 
it is noteworthy that {\it TIM} demonstrates significant improvements in both ZH$\Rightarrow$EN and EN$\Rightarrow$ZH directions. 
This surprising finding suggests that MT-oriented LLMs can serve as valuable fallback systems in cases where the quality of commercial NMT systems is unsatisfactory.

To further support our hypothesis, 
we calculate the COMETkiwi scores of the translations generated by {\it GoogleMT} and {\it TIM} against the source text, selecting a group of sentences where {\it GoogleMT} has higher scores than {\it TIM} by more than 3 points, and another group where {\it TIM} has higher scores than {\it GoogleMT}.
To mitigate the impact of sentence lengths, we retain only those sentences containing fewer than 60 tokens.
Next, we use gpt2-large\footnote{https://huggingface.co/gpt2-large} to calculate the perplexity for the two groups.
The perplexity for sentences in which {\it GoogleMT} excels is 38.61, whereas for sentences in which {\it TIM} performs better, it is 45.51.
The MT-oriented LLM showcases superior proficiency in handling complex source language sentences, as reflected by higher perplexity scores.

\paragraph{Case Study.} 
In Table \ref{tab_case_study}, we provide several examples that are hard for {\it GoogleMT} to handle but are solved well by {\it TIM}. 
It shows that the NMT system struggles to understand the meaning of some professional terms and fails to produce suitable translations for ill-informed text.
In contrast, MT-oriented LLM demonstrates its superiority in handling such issues, which can be attributed to its enhanced ability to comprehend rare, specialized words, and informal texts.

\begin{figure*}[!t]
\centering
\includegraphics[width=1.0\linewidth]{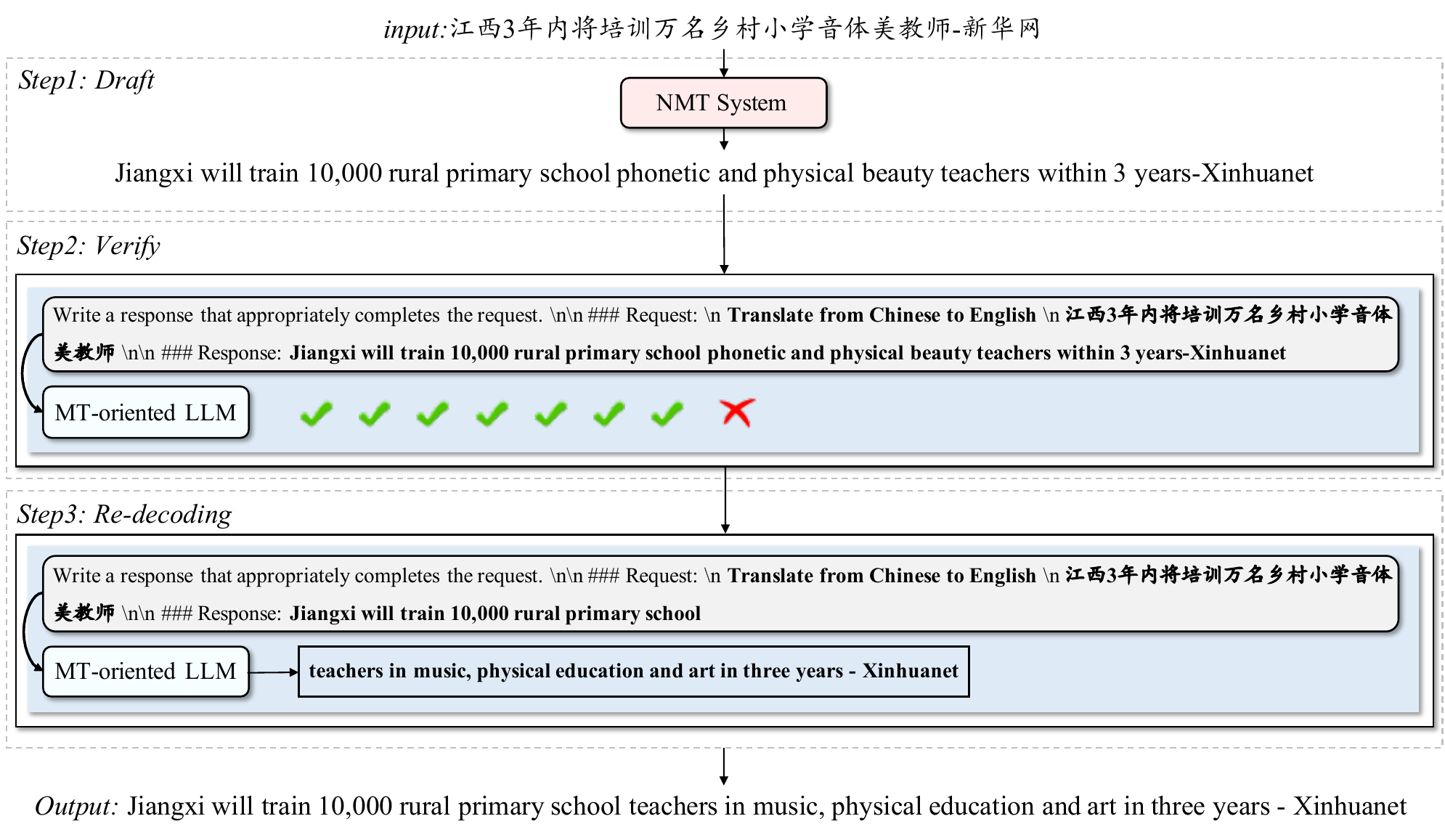}
\caption{
{\bf Cooperative Decoding.} 
The NMT model generates the initial translation (referred to as {\it draft}), and the MT-oriented LLM assesses the quality of the {\it draft} and takes over from the error position, performing verification and re-decoding steps ({\it Verify} and {\it Re-decoding}).
}
\label{fig_cooperative_decoding}
\end{figure*}

\subsection{Discussion}
The analysis in Section \ref{sec_wmt22} demonstrates the accuracy of commercial NMT systems, likely due to their extensive training and cross-attention capability. 
MT-oriented LLMs, known for their paraphrastic nature, can further enhance NMT's ability to handle figurative text translations. 
Moreover, MT-oriented LLMs excel on WebCrawl test sets (Section \ref{sec_webcrawl}), particularly with specialized terminology and ill-formed sentences. This suggests that an effective hybrid framework combining NMT and LLMs can handle challenging input domains and figurative text.
Based on these insights, we propose to investigate an effective hybrid framework 
to answer the second question: {\it How can we effectively harness the capabilities of LLMs to enhance translation quality?}

{Hybrid Threshold} \cite{Hendy} employs the NMT model as the primary translation system, with a quality estimation module (e.g., COMETkiwi) to assess the translation. 
If the quality falls below a certain threshold, the GPT-like model is used as an alternative translation engine. 
However, it faces two main challenges: autoregressive decoding latency and reliable quality estimation.
The practical implementation of the hybrid approaches must ensure efficient decoding and high translation quality.

\section{Cooperative Decoding}




We propose an innovative cooperative decoding approach.
This approach 
leverages an NMT model as a pretranslation model and incorporates an MT-oriented LLM as a quality assessment module and fallback model if needed.
The overview of cooperative decoding is shown in Figure \ref{fig_cooperative_decoding} and we will give a detailed description of each step in the following.

\begin{table*}[!t]
\centering
\small
\setlength{\tabcolsep}{1.5mm}{
\begin{tabular}{lccccc|ccccc}
\toprule
 & \multicolumn{5}{c}{\bf ZH$\Rightarrow$EN} & \multicolumn{5}{c}{\bf EN$\Rightarrow$ZH} \\
 {\bf Method} & {\bf COMET} & {\bf COMETk.} & {\bf Token/s} & {\bf Speedup} & {\bf Ratio} & {\bf COMET} & {\bf COMETk.} & {\bf Token/s} & {\bf Speedup} & {\bf Ratio} \\
\midrule
GoogleMT & 76.8 & 72.8	& - & - & - & 81.9 & 74.5 & - & - & - \\
TIM & 75.6 & 72.3 & 21.8 & 1.0$\times$ & - & 83.0 & {76.0} & 20.7 & 1.0$\times$ & - \\
\midrule
CoDec-4 & 76.7 & 73.0 & 28.5 & 1.3$\times$ & 24.44 & 83.3 & {\bf 76.1} & 24.1 & 1.2$\times$ & 21.64 \\
CoDec-8 & {\bf 77.1} & 73.2 & 32.0 & 1.5$\times$ & 38.83 & {\bf 83.4} & {\bf 76.1} & 25.8 & 1.3$\times$ & 32.69 \\
CoDec-16 & {\bf 77.1} & {\bf 73.3} & 38.7 & 1.8$\times$ & 55.11 & 83.1 & 76.0 & 29.7 & 1.4$\times$ & 46.06 \\
CoDec-32 & {\bf 77.1} & 73.2 & 47.9 & 2.2$\times$ & 67.36 & 83.0 & 75.8 & 33.6 & 1.6$\times$ & 57.06 \\
CoDec-64 & 77.0 & 73.1 & 57.7 & 2.7$\times$ & 76.23 & 82.7 & 75.6 & 38.7 & 1.9$\times$ & 66.25 \\
CoDec-128 & 77.0 & 73.0 & 73.5 & 3.4$\times$ & 84.29 & 82.6 & 75.4 & 45.5 & 2.2$\times$ & 74.35 \\
\bottomrule
\end{tabular}}
\caption{
\label{tab_results_speedup}
{\bf Effect of different values of $k$ (Eq. \ref{eq_k}) for CoDec}.
We present the results on ZH$\Rightarrow$EN and EN$\Rightarrow$ZH including COMET-22, COMETkiwi, decoding speed measured by tokens per second, decoding speedup, and the ratio of the number of tokens accepted at the verification stage to the total tokens of the draft. The choice of $k$ should be considered to strike a balance between performance and efficiency.
}
\end{table*}


\paragraph{Step1: Draft Generation.}
Given an input source sentence $x$, the NMT system generates the translation $o$ using an autoregressive decoding strategy like beam search.
The difference is that the translation $o$ is considered as a draft and requires further confirmation or modification before being used as the final output.

\paragraph{Step2: Verification.}
We feed $o$ into the MT-oriented LLM in a forward process, which fully utilizes parallel computing.
The procedure is the same as training LLMs, and we can obtain a probability distribution $v_t$ at each position, which is modeled as $P(o_t|o_{<t},x)$.
The distribution can be regarded as the confidence of the LLM given the specific prefix of the draft $o$, and we use it to verify the prediction of the NMT model.
One straightforward approach for verification is to check whether the token with the highest probability matches the prediction of NMT.
If $o$ is fortunately exactly the same with $\{{\rm argmax}(v_1),...,{\rm argmax}(v_n)\}$, the inference will finish with $o$ as the final translation.
However, high-quality generation does not follow a distribution of the highest probability of the next tokens, and the tokens in $o$ that can be regarded as accurate may appear outside of the top-1 selection, like in beam search.
To address this issue, 
we relax the matching constraint using the top-{\it k} candidates of the LLM and define the verification criterion as 
\begin{equation}
    o_{t} \in {\rm top-}{k}(v_t). \label{eq_k}
\end{equation}

\paragraph{Step3: Re-decoding.}
The verification is performed from left to right, and we end the verification once there is a situation that does not meet the verification criteria, i.e., $o_{t'}\notin{\rm top-}{k}(v_{t'})$.
Then, we feed the verified prefix $o_{t'-1}$ into the MT-oriented LLM and use it to re-decode the subsequent sequence.
Compared to totally replacing NMT models with MT-oriented LLMs, our cooperative decoding can speed up the whole inference process due to the expensive cost of autoregressive decoding.
The speedup is more significant when the longer draft is accepted.
Moreover, the cooperative mechanism alleviates the issue of inaccuracy of LLMs by exploiting the output of NMT models. 

\section{Experiments}

\subsection{Main Results}

We merge the WMT22 and WebCrawl test sets to simulate the distribution of translation requests in real-world scenarios.
For CoDec, we use GoogleMT as the NMT system, and TIM as the MT-oriented LLM. 
In particular, we set the threshold as the 50th percentile of COMETkiwi scores of GoogleMT \citep{Hendy}.
We use the MT-oriented LLM to generate the translation only when the COMETkiwi score of the NMT translation is under the threshold.
We use beam search with a beam size of 4 for TIM during inference.
The decoding and speed measurement processes are performed on a single A100 GPU.

\paragraph{Effect of different values of $k$.}

Intuitively, as $k$ increases, cooperative decoding can accept a wider range of tokens in NMT translations during the verification stage. 
As a result, less content needs to be re-decoded by LLMs, leading to a reduction in processing time. 
Here, we examine the performance of CoDec under various values of $k$.

As shown in Table \ref{tab_results_speedup}, with the increase of $k$, the ratio of tokens accepted on average and the decoding speed increase consistently.
With a larger $k$, {\it CoDeC-128} achieves a 3.4x and 2.2x speedup over {\it TIM} in ZH$\Leftrightarrow$EN. 
This signifies that CoDec effectively reduces decoding latency while maintaining translation quality.
Besides, our CoDec-(*) models exhibit superior performance compared to both {\it GoogleMT} and {\it TIM}.
This highlights the potential of cooperative decoding in improving translation accuracy and overall system performance.
Moreover, models with lower values of $k$, such as {\it CoDec-8}, achieve better translation quality, suggesting that the choice of $k$ should be considered to strike a balance between performance and efficiency.

\paragraph{CoDec vs. Hybrid Threshold.}

In our comparison between CoDec and Hybrid Threshold, 
we utilize different Quality Estimation (QE) methods, including
{\it HT(Random)}, where 50\% of GoogleMT's translations are randomly replaced with TIM's translations,
{\it HT(BLEURT-12)}, which uses BLEURT-20-D12\footnote{https://huggingface.co/lucadiliello/BLEURT-20-D12} as the QE method; 
{\it HT(BLEURT-20)}, which employs BLEURT-20\footnote{https://huggingface.co/lucadiliello/BLEURT-20} as the QE method; 
and {\it HT(COMETk.)}.
Additionally, CoDec is integrated into the Hybrid Threshold pipeline as a comparative system, referred to as {\it HT(COMETk.) w/ CoDec}. 
Furthermore, we follow \citet{Hendy} to use Hybrid Max-Routing to establish an upper bound by selecting the best translation from either system based on the COMETkiwi.

\begin{table}[t]
\centering
\small
\setlength{\tabcolsep}{0.5mm}{
\begin{tabular}{lcc}
\toprule
\multirow{2}{*}{\bf Model} & {\bf ZH$\Rightarrow$EN} & {\bf EN$\Rightarrow$ZH} \\
& COMET/COMETk. & COMET/COMETk. \\ 
\midrule
GoogleMT & 76.8/72.8 & 81.9/74.5 \\
TIM & 75.6/72.3 & 83.0/76.0 \\
\midrule
HT(Random) & 76.2/72.5 & 82.4/75.2 \\
HT(BLEURT-12) & 76.3/72.8 & 82.6/75.1 \\
HT(BLEURT-20) & 76.3/72.8 & 82.7/75.2 \\
HT(COMETk.) & 76.5/73.1 & {83.3}/{\bf 76.2} \\
\ \ \ \ w/ CoDec & {\bf 77.1}/{\bf 73.3} & {\bf 83.4}/{\bf 76.2} \\
\midrule
CoDec-8 & {\bf 77.1}/{73.2} & {\bf 83.4}/76.1 \\
\midrule
Max-Routing & 77.4/74.3 & 84.0/76.5 \\
\bottomrule
\end{tabular}}
\caption{
{\bf Comparison among CoDec-8 and Hybrid Threshold with different QE methods}. 
Different QE methods in Hybrid Threshold (HT) show varying performances, whereas CoDec surpasses most HT models. Our CoDec achieves a better balance between efficiency and effectiveness.
\label{tab_compared_with_ht}
}
\end{table}

The performance comparison in Table \ref{tab_compared_with_ht} reveals a notable performance disparity between {\it GoogleMT} and {\it Max-Routing}. 
This result supports our assertion that MT-oriented LLMs can play a crucial role as reliable fallback systems for NMT systems.
Moreover, the different QE modules employed in Hybrid Threshold yield varying performances, highlighting the dependence of Hybrid Threshold's performance on the precision of the QE modules and the quality of LLM translations used as replacements. 
In contrast, {\it CoDec-8} surpasses most of the Hybrid Threshold models and achieves competitive results with {\it HT(COMETk.) w/ CoDec}, suggesting that the QE modules may not be necessary. 
The findings validate that our approach achieves a better balance between efficiency and effectiveness, resulting in enhanced translation quality without compromising system efficiency.

\subsection{Human Evaluation}
In addition, we carry out a human evaluation on the WebCrawl EN$\Rightarrow$ZH dataset. 
A total of 300 sentences are randomly selected from the test set, and two individuals are asked to evaluate the translations produced by GoogleMT, HT(COMETk.), and our CoDec-8. 
We use the commonly used pairwise comparison method to count the number of better, similar, and worse translations from System 1 rather than System 2.
The result of CoDec vs. GoogleMT is 144:115:41, while the result of CoDec vs. HT(COMETk.) is 106:130:64.
It shows that our CoDec significantly outperforms the commercial NMT system and performs better than the Hybrid Threshold without an additional quality evaluation module.

\begin{table}[t]
\centering
\small
\setlength{\tabcolsep}{0.7mm}{
\begin{tabular}{lcccc}
\toprule
\multirow{2}{*}{\bf Model} & \multicolumn{2}{c}{\bf DE$\Rightarrow$EN} & \multicolumn{2}{c}{\bf ZH$\Rightarrow$EN} \\
& COMET/ChrF & Suc. & COMET/ChrF & Suc. \\ 
\midrule
Lingua Custodia & 73.5/61.8 & 62.2 & 60.9/32.6 & 74.7  \\
$\text{UEDIN}_{\rm LLM}$ & 81.3/60.0 & 58.8 & \textbf{75.7}/\textbf{41.2} & 75.3 \\
\midrule
GoogleMT & 80.3/54.3 & 55.0 & 75.3/41.0 & 67.1 \\
TIM w/o term & 79.6/54.0 & 54.1 & 73.8/38.5 & 58.6 \\
TIM w/ term & {\bf 82.3}/65.2 & {\bf 82.5} & 73.4/39.4 & {\bf 85.0} \\
\midrule
CoDec-8 & 80.7/56.1 & 59.0 & 75.3/41.0 & 76.4 \\
\bottomrule
\end{tabular}}
\caption{
{\bf Performance on WMT23 terminology translation}.
``Suc.'' denotes Terminology Success Rate.
Our CoDec combines NMT's superior translation quality with the constrained translation capabilities of MT-oriented LLMs.
\label{fig_terminology}
}
\end{table}

\subsection{Terminology Translation}
Unlike conventional NMT models, MT-oriented LLMs enable them to exploit instructions to handle various translation scenarios.
Here, we apply CoDec to assess the effectiveness of incorporating instructions in a dedicated terminology translation test set obtained from WMT23\footnote{https://wmt-terminology-task.github.io/}.
The result is shown in Table \ref{fig_terminology}, evaluated by COMET, ChrF, and Terminology Success Rate.
The data statistics and details of baselines can be found in Appendix \ref{appendix_terminology}.

The results indicate that the use of terminology information in instructions, as demonstrated by {\it TIM w/ term}, enables MT-oriented LLMs to achieve constrained machine translation, resulting in more accurate domain-specific terminology in the translated output.
When compared to {\it Lingua Custodia} and {\it UEDIN}$_{\rm LLM}$ \cite{semenov-etal-2023-findings}, {\it CoDec-8} combines the advantages of higher translation quality offered by NMT and the constrained translation capabilities of MT-oriented LLMs. 
This combination leads to higher-quality translations while maintaining a higher Terminology Success Rate.

\section{Conclusion}


We explore the strengths of both NMT and LLM and propose CoDec that integrates the two to achieve superior performance compared to existing hybrid frameworks.
Notably, our CoDec offers reduced decoding latency compared to relying solely on LLMs for inference, and it does not require any modifications to the target LLMs.
We believe that exploring more effective utilization of LLMs while considering practicality in both industry and academia is a valuable direction.

\section{Limitations}
This paper primarily concentrates on enhancing translation performance for medium and high-resource language pairs. Further investigation is required to analyze the translation characteristics of different systems in low-resource languages, which we defer to future research.

Additionally, the draft translations were validated by directly utilizing the top-$k$ candidates predicted by the target MT-oriented LLM. We acknowledge that the implementation of more meticulously designed token-level validation methods has the potential to further enhance CoDec, and we consider it as an avenue for future exploration.


\bibliography{anthology,custom}

\appendix


\section{WMT22 test sets} \label{appendix_wmt22}
To prevent data leakage \citep{arxiv2023:unreasonable}, we analyze the WMT22 test sets, consisting of recent content from diverse domains including news, social media, e-commerce, and conversation. 
The test sets consist of the following number of samples for each language pair: 
German-to-English (DE$\Rightarrow$DE) - 1984 samples, English-to-German (EN$\Rightarrow$DE) - 2037 samples, Chinese-to-English (ZH$\Rightarrow$EN) - 1875 samples, English-to-Chinese (EN$\Rightarrow$ZH) - 2037 samples, Russian-to-English (RU$\Rightarrow$EN) - 2016 samples, English-to-Russian (EN$\Rightarrow$RU) - 2037 samples, Japanese-to-English (JA$\Rightarrow$EN) - 2008 samples, English-to-Japanese (EN$\Rightarrow$JA) - 2037 samples.

\section{Unaligned Source/Target Words.}\label{appendix_usw}
For English and German, we utilize the Moses tokenizer\footnote{https://github.com/moses-smt/mosesdecoder/tree/master/scripts/tokenizer}. 
We use jieba\footnote{https://github.com/fxsjy/jieba} and MeCab\footnote{https://github.com/SamuraiT/mecab-python3} for Chinese and Japanese, respectively.
We use {\it awesome-align}\footnote{https://github.com/neulab/awesome-align} \citep{dou-neubig-2021-word} to obtain the word alignments.
Unaligned source words (USW) indicate the number of words in the source text that have no corresponding translation in the target sentence.
Unaligned target words (UTW) assess the degree to which words are potentially added or inserted into the translation without any basis or support from the source sentence.


\section{WebCrawl test sets}

\label{appendix_webcrawl}
To acquire the data, we follow the process outlined below:
\begin{itemize}
    \item We extract snippets from web pages and use an in-house sentence segmentation tool to split them into individual sentences.
    \item We employ sensitive word filters, language identification tools, length ratio checks, and perplexity scores to filter out sentences of lower quality. 
    \item We utilize Google Translator to obtain translations of the sentences, with a primary focus on the Chinese$\Leftrightarrow$English directions.
    \item We calculate COMETkiwi scores and retain sentences with scores below 65.
\end{itemize}

In this way, we collected a total of 889 Chinese sentences and 1195 English sentences as our final test set, named {\it WebCrawl test sets}. 
We hire 2 annotators who have degrees in English Linguistics to annotate translations. 
Before formal annotation, annotators were asked to annotate 100 samples randomly extracted from the dataset, and based on average annotation time we set a fair salary (i.e., 30 dollars per hour) for them.


\begin{table*}[!h]
\centering
\small
\setlength{\tabcolsep}{2mm}{
\begin{tabular}{l|cccc|cccc}
\toprule
{\bf System} & {\bf ChrF} & {\bf SacreBLEU} & {\bf ChrF} & {\bf SacreBLEU} & {\bf ChrF} & {\bf SacreBLEU} & {\bf ChrF} & {\bf SacreBLEU} \\
\midrule
& \multicolumn{2}{c}{\it DE$\Rightarrow$EN} & \multicolumn{2}{c|}{\it EN$\Rightarrow$DE} & \multicolumn{2}{c}{\it ZH$\Rightarrow$EN} & \multicolumn{2}{c}{\it EN$\Rightarrow$ZH} \\
WMT-Best & 58.5 & 33.4 & 64.6 & {\bf 38.4} & {\bf 61.1} & {\bf 33.5} & 41.1 & 44.8 \\
\hdashline[0.5pt/0.5pt]
GoogleMT & {\bf 59.1} & {\bf 34.1} & {\bf 64.7} & 37.5 & 60.0 & 29.4 & {\bf 45.8} & {\bf 50.5} \\
MicroMT & 58.8 & 33.9 & 64.7 & 37.5 & 60.0 & 29.4 & {\bf 45.8} & {\bf 50.5} \\
\hdashline[0.5pt/0.5pt]
BayLing-7B & 53.6 & 28.2 & 53.6 & 25.7 & 49.9 & 20.3 & 34.5 & 38.2 \\
TIM-13B & 56.9 & 31.7 & 60.8 & 33.2 & 56.8 & 26.9 & 42.4 & 46.9 \\
\midrule
& \multicolumn{2}{c}{\it RU$\Rightarrow$EN} & \multicolumn{2}{c|}{\it EN$\Rightarrow$RU} & \multicolumn{2}{c}{\it JA$\Rightarrow$EN} & \multicolumn{2}{c}{\it EN$\Rightarrow$JA} \\
WMT-Best & 68.9 & 45.1 & 58.3 & 32.4 & 49.8 & 24.8 & 36.8 & 27.6 \\
\hdashline[0.5pt/0.5pt]
GoogleMT & {\bf 69.1} & {\bf 45.7} & 59.5 & 34.3 & {\bf 51.8} & {\bf 26.2} & {\bf 37.6} & {\bf 28.2}  \\
MicroMT & {\bf 69.1} & {\bf 45.7} & {\bf 59.6} & {\bf 34.9} & 49.5 & 24.6 & 34.8 & 25.1  \\
\hdashline[0.5pt/0.5pt]
BayLing-7B & 60.4 & 34.7 & 35.5 & 14.8 & 34.7 & 11.6 & 9.6 & 4.5 \\
TIM-13B & 65.7 & 40.4 & 54.6 & 28.5 & 46.3 & 21.6 & 29.6 & 19.7 \\
\bottomrule
\end{tabular}}
\caption{
\label{tab_results_chrf_bleu}
{Experimental results on the WMT22 test sets}. 
}
\end{table*}

\begin{figure*}[!h]
\centering
\includegraphics[width=1.0\linewidth]{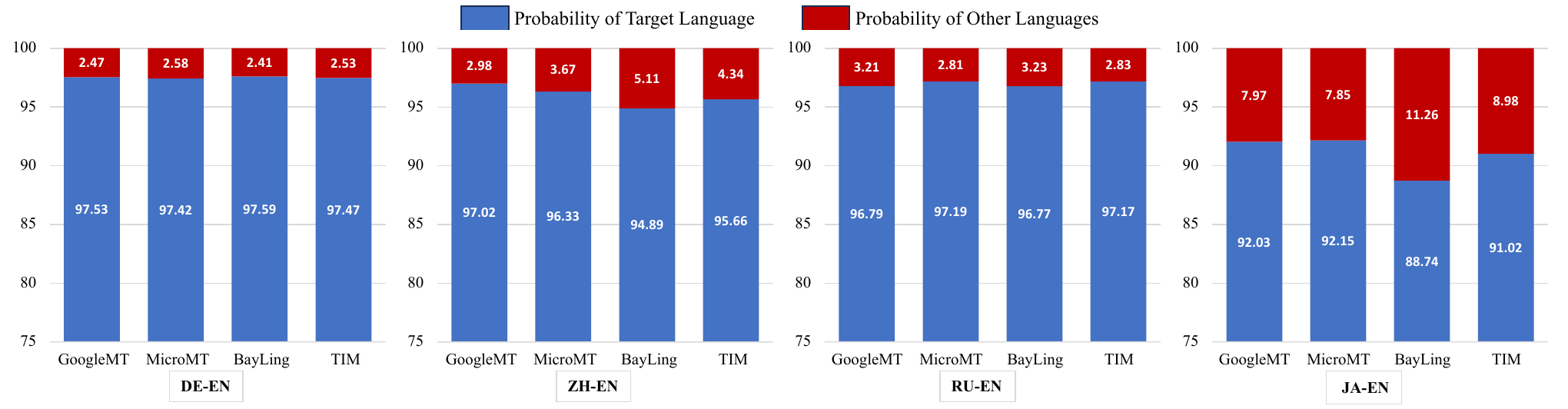}
\caption{
{\bf Off-target rates (\%) of translations.} MT-oriented LLMs exhibit a higher prevalence of off-target translations than NMT systems.
}
\label{fig_lang_prob}
\end{figure*}


\section{Terminology Translation}
\label{appendix_terminology}
Terminology translation is an extensively encountered application scenario, where the NMT (Neural Machine Translation) model is expected to precisely handle the provided domain-specific terminology.
In this experiment, we use the prompt ``\texttt{\{srcWord\} means \{tgtWord\}. Translate the following sentences from \{src\} to \{tgt\}, and muse use the given word translations.\{line\}}'' for inference of TIM.
The numbers of sentences on Zh$\Rightarrow$En and De$\Rightarrow$En are 2640 and 2963, respectively.
The average numbers of terms per segment on Zh$\Rightarrow$En and De$\Rightarrow$En are 3.8 and 1.1, respectively.
We only highlight a few systems that achieved the best performance on specific metrics in the competition findings \cite{semenov-etal-2023-findings}.
Lingua Custodia, which utilizes a specialized Transformer architecture to ensure the inclusion of given terminology in the translation. Additionally, the $\text{UEDIN}_{\rm LLM}$ employs ChatGPT with prompts specifically designed for terminology translation.

\section{Different from traditional NMT with additional language models
}
Traditional language models, such as causal language models are usually used as decoder initialization or reranking to improve fluency. We do not consider the prediction probabilities of LLMs during the decoding process of NMT. 
Instead, we treat LLMs as independent translation systems and introduce speculative execution as a fusion approach for NMT systems and MT-oriented LLMs. 
Notably, we do not introduce additional monolingual training data and our research does not encompass back-translation.
Building upon this insight, we propose a hybrid framework that combines the strengths of both NMT systems and MT-oriented LLMs and alleviates the limitations of individual systems.

\begin{table*}[!h]
\centering
\small
\begin{tabular}{ll}
\toprule
{\bf System} & {\bf Translation} \\
\midrule
\multicolumn{2}{c}{\it Terminology/abbreviations} \\
Source & Art. 18 GDPR: Right to restriction of data processing if the requirements Art. 18 para 1 lit. a to d are fulfilled.  \\
GoogleMT & 艺术。 GDPR 第 18 条：如果满足第 18 条的要求，则有权限制数据处理。 18 段 1 字。 a到d均满足。  \\
TIM & 《通用数据保护条例》第 18 条：如果满足第 18 条第 1 款 a 至 d 项的要求，则有权限制数据处理。 \\
\midrule
\multicolumn{2}{c}{\it Ill-informed text} \\
Source & 批《道路机动车辆生产企业及产品公告》中，江淮 \\
GoogleMT & In the batch of "Announcement of Road Motor Vehicle Manufacturers and Products", JAC \\
TIM & In the "Road Motor Vehicle Manufacturers and Products Announcement", Jianghuai \\
\midrule
\multicolumn{2}{c}{\it Complex，Repetition-containing} \\
Source & let mut v = vec![10, 20, 30];       let handle = thread::spawn(|| {    v.push(10);       }); \\
GoogleMT & 让 mut v = vec![10, 20, 30];让句柄 = thread::spawn(|| { v.push(10); }); \\
TIM & let mut v = vec! [10,20,30]; let handle = thread::spawn (|| { v.push (10); }); \\
\bottomrule
\end{tabular}
\caption{
{\bf Case Study}. We present examples of several translation challenges that pose difficulties for NMT systems but are effectively mitigated by MT-oriented LLMs.
}
\label{appendix_tab_case_study}
\end{table*}

\section{About speedup}
The time consumption of the Hybrid Threshold is the sum of the inference time for both the NMT systems and the MT-oriented LLM, whereas the CoDec requires only the inference time of the NMT systems and a small amount of calculation of the LLM. 
Considering the relatively negligible time consumption of Google Translate, we did not specifically factor in its inference time in our analysis, as it does not significantly impact the overall performance comparison.

\end{CJK}
\end{document}